%% file: 2019_IMPI_BrainTumorSeg_MissingChannels_Final.tex
\RequirePackage{amsmath}
\documentclass[runningheads,a4paper]{llncs}

\usepackage{amssymb}
\usepackage{graphicx}
\usepackage{subfigure}
\usepackage{url}
\urldef{\mailsa}\path|{yshen22,|
\urldef{\mailsb}\path|mgao8}@buffalo.edu|    
\newcommand{\keywords}[1]{\par\addvspace\baselineskip
\noindent\keywordname\enspace\ignorespaces#1}

\usepackage{xcolor}

\newcommand{\etal}{\textit{et al}. }

\begin{document}

\mainmatter  

\title{Brain Tumor Segmentation on MRI with Missing Modalities}

\titlerunning{Brain Tumor Segmentation on MRI with Missing Modalities}

%
%
\author{ 
Yan Shen\and Mingchen Gao
}
\authorrunning{Y. Shen \etal}

\institute{
Department of Computer Science and Engineering,
University at Buffalo,\\
The State University of New York, Buffalo, USA\\
\mailsa\mailsb\\
}

%
%

\toctitle{Brain Tumor Segmentation on MRI with Missing Modalities}
\tocauthor{Authors' Instructions}
\maketitle

\begin{abstract}


Brain Tumor Segmentation from magnetic resonance imaging (MRI) is a critical technique for early diagnosis. However, rather than having complete four modalities as in BraTS dataset, it is common to have missing modalities in clinical scenarios. We design a brain tumor segmentation algorithm that is robust to the absence of any modality. Our network includes a channel-independent encoding path and a feature-fusion decoding path. We use self-supervised training through channel dropout and also propose a novel domain adaptation method on feature maps to recover the information from the missing channel. Our results demonstrate that the quality of the segmentation depends on which modality is missing. Furthermore, we also discuss and visualize the contribution of each modality to the segmentation results. Their contributions are along well with the expert screening routine.

\keywords{Brain tumor segmentation, Multi-modality, Domain adaptation, Self-supervised learning
}
\end{abstract}

\input{./subtex/intor1}

\input{./subtex/method1}
\input{./subtex/experiment1}
\input{./subtex/conclusion1}

\bibliographystyle{abbrv}
\bibliography{IPMI2019}

\end{document}

%% file: subtex/intor1.tex
\section{Introduction}
In the United States, it is estimated that 25,000 new patients are diagnosed with brain cancer every year. While average five-year survival rate is just above one in three, early diagnosis is important to increase life quality of patients. MR imaging are the most common kinds of screening methods on brain pathology. MR images provide some primary indicators on unhealthy tissue matters, and have great impact on improved diagnosis, tumor classification and treatment planning.

While brain tumor segmentation requires professional knowledge to distinguish unhealthy tissues from healthy ones. Those tasks are expensive and limited. Computer-aided automatic segmentation tools provide an important reference for diagnosis. The automatic input-output pipeline can provide patients instant diagnosis that gives them the opportunity to have first-aid on what suspects to be a pathology.

The success of deep learning has made computer program an indispensable aid to physicians for analyzing on brain tumor development \cite{hoo2016deep}. Such automatic detection methods based on multi-layer neural networks have been widely applied to segmentation on brain tumors. MICCAI Brain Tumor Segmentation (BraTS) challenge collected so far the largest collection of brain tumor segmentation dataset that is publicly available. They use well-defined training and testing splits, thereby allowing us fair comparison among different approaches. Mohammad \etal are the first to apply deep neural network methods to segment brain tumors on BraTS dataset \cite{havaei2017brain}. They use Fully Convolutional Networks (FCN) to perform pixel level classification by taking local patches as input. Their method becomes the baseline for deep learning based approach on BrainTumor Segmentation. In our knowledge, the best result on Brain Tumor Segmentation is produced by RA-UNet \cite{jin2018ra} which proposes a 3D hybrid residual attention-aware segmentation method. They achieve Dice score of 0.8863 on whole tumor segmentation. Currently, state-of-art result on BraTS dataset can even produce comparable results with human experts.
 
However current state-of-art brain tumor segmentation benchmark requires complete MR images in all T1, T2, T1c and Flair modalities, as provided in the dataset. Multi-modality images provide complementary information to distinguish tissues and anatomies. Segmenting brain tumors from MR images learns a complicated function on all four modalities parameterized both by anatomical morphology and local pixels values.  However, in practical situations, different hospitals may follow different protocols and procedures when performing MR images. Missing one of the four modalities is common. This may pose some challenges on simultaneous available requirements of four complete modalities on MR images.

\paragraph{\bf Related Work} Image segmentation a task of clustering pixels into salient image regions corresponding to their characteristics. They typical generate labels on the pixel level. Earlier time image segmentation uses handcrafted filter banks like Gaussian Kernels, Fourier Transformations, Wavelets to pre-process the image and later trains a SVM for classification. Succeeding work \cite{zhang2001segmentation} uses statistical approach like CRF to infer a fine boundary. CRF approaches incorporates local evidence in unary assignments and models interactions between label assignment in a probabilistic graphical model. In the era of deep learning, convolutional neural networks (CNNs) are considered as the state-of-the-art in biomedical image segmentation. Deep approaches include a forward pipeline to inference target from inputs and a backward pipeline to learn network parameters. Early deep learning method on image segmentation uses FCN classification networks that perform pixels level classifications by taking local region inputs. Later approaches turn to U-Net \cite{ronneberger2015u} structure that include a down-sampling path to encode input images and an up-sampling path to decode feature maps to segmentation map. Further work \cite{zheng2015conditional} adds a RNN layer like a CRF to further refine the boundary of segmentation maps produced by U-Nets. 

Deep Learning approaches for brain tumor segmentation include end-to-end U-Net \cite{kao2018brain,isensee2017brain}, regional proposals \cite{wang2017automatic}, hierarchy classification \cite{chen2018mri} and level set \cite{le2018deep}. Hierarchical models have been proposed to firstly classify pixels into background and whole tumor using T2 and Flair channels, the whole tumor is then further classified into four classes based on all four modalities \cite{chen2018mri}.

 There are also a number of methods proposed on dealing with missing data in medical imaging. Primitive methods include building separate models for each condition. However each model could only fit a specific combination of input modalities. Domain adaptation methods are widely used to learn a unified model that could apply for a wide spectrum of channel modes. Ganin \etal \cite{ganin2015unsupervised} use adversarial loss on intermediate feature maps of two domains and thus transfer the network learnt in one domain to the other. Manders \etal \cite{manders2018simple} use a class specific adversarial loss on feature maps to transfer the learnt network from source domain to target domain. Generative model is also proposed by Zhang \etal \cite{zhang2018deep} to synthesize missing channel from available channels. Havaei \etal \cite{havaei2016hemis} builds a unified model from purely self-supervised training pipeline for each individual channel. And they made combined prediction on multi-channels by merging the feature maps from each channels and computing the mean and variance.
 
\paragraph{\bf Contributions} 

In our paper, we use self-supervised learning by randomly removing one modality during training to enhance the model to handle the instances with a missing modality. While trained with a missing modality, we use an extra adversarial loss to ensure the modal generate similar features as in full modality situation. Rather than directly adapting one domain to another in previous methods, we consider domain diversities from different channels. We take the combined features from all the domains and adapt it to encode distribution on any of remaining channels after dropping one.

Our contributions in this paper are three folds:
\begin{itemize}
\item Firstly, we learn a network that can be broadly applied to tolerate on missing any modality without the need of fitting to specific combinations of remaining modalities. 

\item Secondly, we put forward a novel domain adaptation model. Our adaptation model uses adversarial loss to adapt feature maps from missing modalities to the one from full modalities.

\item Finally, we decouple the segmentation contribution to each individual modality and thus generate on interpretations of our segmentation on four modalities of MR images.     
\end{itemize}

%% file: subtex/method1.tex
\section{Training Modality Missing Tolerant Network: a joint objective of missing modality reconstruction and domain adaptation}
\subsection{U-Net Structure: Modality-Separate Encoding Path and Feature Fusion at Various Resolutions on Decoding Path}
\label{subsection:fullChannel}

\begin{figure}[htb]
	\centering
	\vspace{-0.5cm}
	\includegraphics[width=\linewidth]{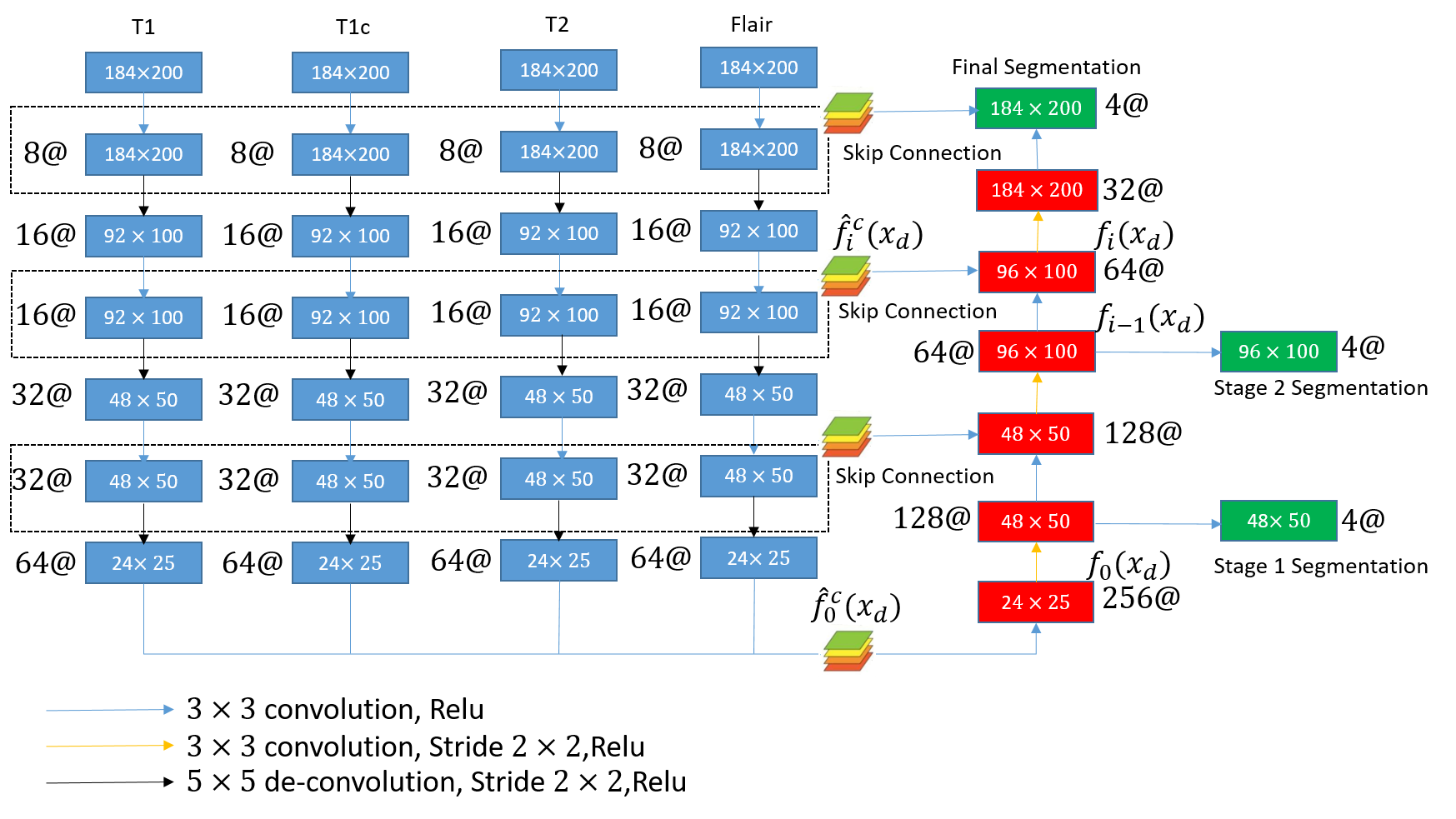}
	\caption{The architecture of our 4 channel input segmentation network} \label{fig:arch}
\end{figure}

Our basic model architecture follows U-net, as shown in Fig. \ref{fig:arch}. Each modality is fed into the model as one channel. Blue box indicates separated encoding path of the multi-channel MR image input. Red box indicates decoding path following the encoding of multi-channel input on various levels of feature maps. Green box indicates multi-stage segmentation outputs. Encoding path includes multiple feed-forward down-sampling convolutional layers after each channel input. And it ends up with 64 features map of resolution $24\times25$ from each input channel. Individual encoding path of each channel have independent sets of parameters and have no connection with each other before decoding. Though cross-channel filter banks could be used to extract features by contrasting different input channels at early stage in a more straight-forward way, our separated feature maps on four different channels can be used to encode abstractions of segmentation information that is tolerant on channel loss. Our separately encoded feature maps are fused and decoded in our following decoding path. 

In our network's decoding path, higher resolutions of segmentation outputs are created by successively up-sampling in a bottom-up approach and mixing channel-separate feature maps at various levels. Up-sampling operations use de-convolution operators. Channel-separate feature maps created at encoding path are fused by convolution operators. The features maps at the end of encoding path from each input channel are fused to create bottle-neck feature maps at the same resolution. Two levels of inter-mediate feature maps from each input channel are fused with the feature maps created along the decoding path. We fuse intermediate feature maps from encoding path with the ones from decoding to create skip connections. Our decoding path includes a total of three stages of successive mixing and up-sampling. After each stage, we also generate a segmentation at the same level of the same resolution by convolutions. The final segmentation outputs with the same resolutions as input MR images are generated after the final stage.

We train on the lower resolution segmentation outputs at the intermediate feature maps to force our network learn a hierarchical representation of segmentation related information from coarse to fine, where the structural information is encoded at the lower resolution feature maps and the boundary level information is encoded at the higher resolution feature maps. In this way, we generate our segmentation outputs by sequentially inferring area of pathology structures and exact boundaries.

\subsection{Segmentation on Missing Modality Inputs}
\label{subsection:missingChannel}

We first define how our network infers segmentation outputs from missing channel inputs. Formally, let's denote $x_d$ as a sample taken from the training set $\mathcal{D}$, $\mathcal{C}$ as the set of all channels in MR images. And we suppose only channels in set $\mathcal{C}^-_d \in \mathcal{C}$ are available. As we mentioned previously, our U-Net's decoding path mixes the channel-specific feature maps with the feature maps created along the decoding path and then use convolution operations to predict the probability of segmentation at various resolutions. In the condition of missing channel inputs, we set the encoding path of missing channel to zero, and keep the rest the same. For the decoding path, the feature maps are calculated as :

\begin{equation}
       f_{i}(x_d) =  \sum_{c \in {C}^-_d}\hat{W}_i^{c}(\hat{f}_{i}^c(x_d)) + W_{i}(f_{i-1}(x_d)),  
\end{equation}
where $\hat{W}_i^{c}(\cdot)$ denotes the skip convolution kernel on the same level encoding feature map $\hat{f}_{i}^c(x_d)$ from channel c at layer $i$, $W^{i}(\cdot)$ denotes the feed-forward convolution kernel on feature map $f_{i-1}(x_d)$ produced at the layer $(i-1)$ before along decoding path. The skip convolution feature maps $\hat{f}_{i}^c(x_d)$ comes from the encoding path after a $3\times 3$ stride 1 convolution on feature map of the same size as $f_{i-1}(x_d)$. At the bottleneck layer $0$, their feature maps only come from convolutions on channel-separated feature maps from encoding layers. For missing channel inputs, we only convolute on those feature maps from available channels and zero out the rest.
\begin{equation}
f_{0}^-(x_d) =  \sum_{c \in {C}^-_d}W_0^{c}(\hat{f}_{0}^c(x_d))   
\end{equation}
where $W_0^{c}(\cdot)$ denotes the convolution kernel on the feature map $\hat{f}_{0}^c(x_d)$ from channel c. The rest available channels for encoding path keep the same as in the architecture of full channels in described in \ref{subsection:fullChannel}.
\begin{figure}[htb]
	\centering
	\vspace{-0.6cm}
	\includegraphics[width=\linewidth]{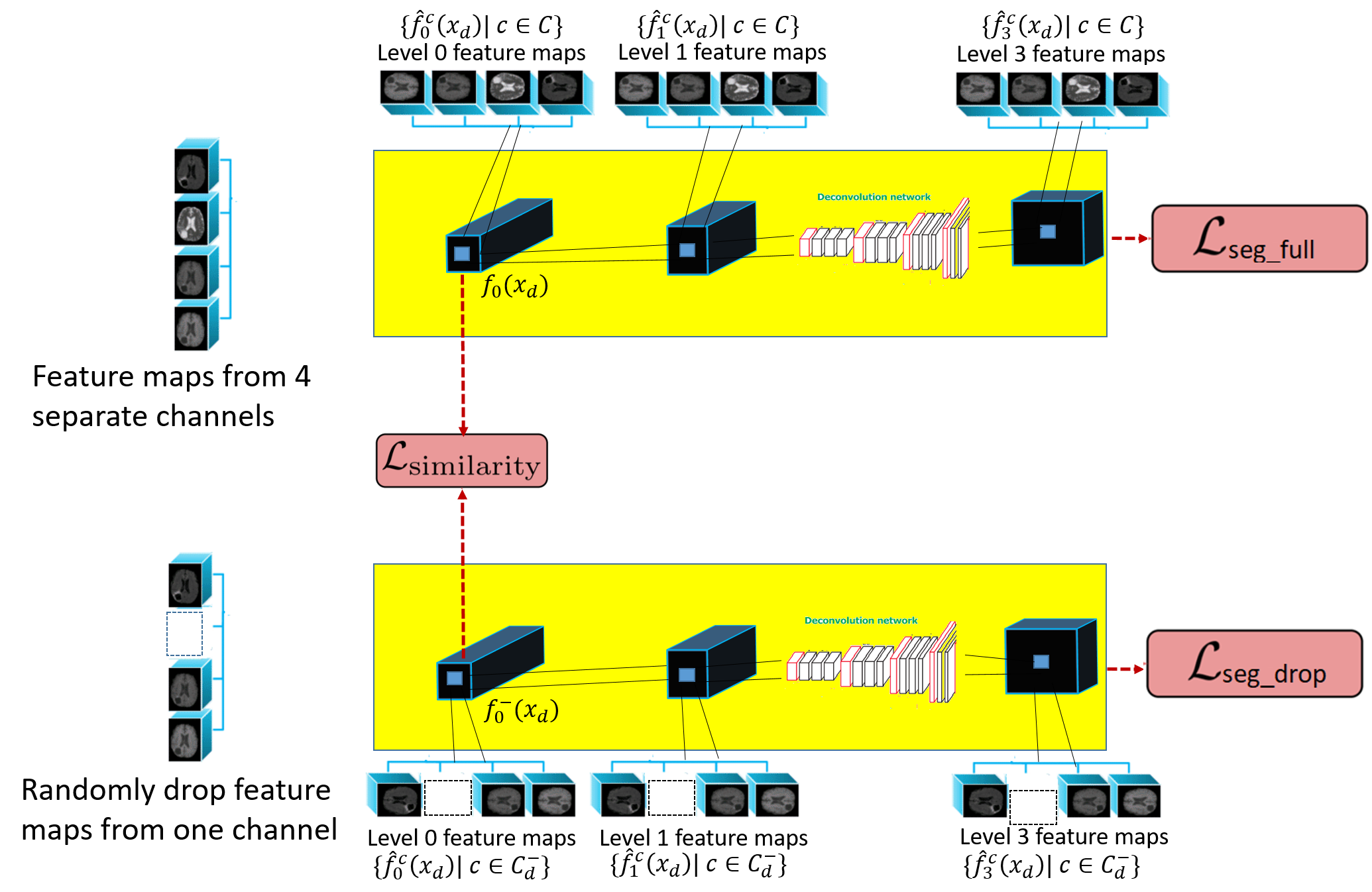}
	\caption{Training of our brain separation networks on missing channel. Encoded feature maps from 4 separate channels are connected to decoding path in full channel input. Encoded feature maps from residue channels are connected on decoding path in missing channel input. We randomly drop one channel in a training step. Our training objective function include the reconstruction loss on full channel input and missing channel input. We also add the similarity loss on decoded feature maps from full channel input and the one from missing channel input.} \label{fig:train}
\end{figure}

\subsection{Training Missing Modality Tolerant Model by Random Modality-Drop}

The model proposed in \ref{subsection:missingChannel} is able to train and test images with missing channels. However, the missing channels will certainly jeopardize the segmentation performance. To make our model robust to channel losses, we initiate a process to compensate the missed modalities. We train the model by randomly dropping out an input channel. The model has independent encoding path has two benefits: (i) to learn a unique feature maps for each input channel (ii) to avoid the false co-adaptations between modalities. Our key idea is to train a model capable of producing meaningful segmentations from any arbitrary combination of available channels.

The objective function is as follows: 
\begin{equation}
\mathcal{L}_{seg\_drop} = \sum_{k=1}^3 \sum_{d \in \mathcal{D}} {^{k}\mathbf{y}_d} \log p_Y( ^{k}\hat{\mathbf{y}}_d |x_d^{(\mathcal{C}^-_d)}), \label{eq:bottleneck}
\end{equation}
where $x_d^{(\mathcal{C}^-_d)}$ is the $|\mathcal{C}|-1$ available channel inputs from $x_d$ after we randomly drop one, ${^{k}\mathbf{y}_d}$ is the k's stage resolution of one-hot encoding of segmentation labels, $p_Y( ^{k}\hat{\mathbf{y}}_d |x_d^{(\mathcal{C}^-_d)})$ is the softmax segmentation prediction. We use a uniform distribution to seed the index of the channel to be dropped at each training step.

\subsection{Domain Adversarial Similarity Loss on Bottleneck Feature Maps}
The domain adversarial similarity loss \cite{ganin2015unsupervised} is used with a discriminator to adapt representations from different domains to a similar distribution. Our whole training diagram is shown in Fig. \ref{fig:train}. We further introduce a domain adversarial similarity loss term on bottleneck feature maps for co-adaptations among abstract representations from different channels. We choose bottleneck feature maps for adaptation because it is where individually predicted abstract features are fused to produce a joint inference on high level representations for segmentations (mostly at structural level).

The outputs at bottleneck layer are the sum of convolutions from all contributing channels as shown in eq. \ref{eq:bottleneck} . In the case of modality-drop, the outputs at bottleneck layer are the sum of convolutions from available channels.

The expectation of $f^-_{0}(x_d)$ could be written as:
\begin{align*}
 \mathbb{E}[f^-_{0}(x_d)] & =  \mathbb{E}_{\delta, x_d}[\sum_{c \in {C}^-_d} W_0^{c}(\hat{f}_{0}^c(x_d)) ]\\
 & = \mathbb{E}_{\delta, x_d}[\sum_{c \in {C}}\delta^{(c)} W_0^{c}(\hat{f}_{0}^c(x_d)) ] \\
& = \mathbb{E}_{\delta|x_d}  [\sum_{c \in {C}}\delta^{(c)}\mathbb{E}_{x_d}[W_0^{c}(\hat{f}_{0}^c(x_d))]]   \\
& = \sum_{c \in \mathcal{C}} p(\delta^{(c)} \neq 0)\mathbb{E}[W_0^{c}(\hat{f}_{0}^c(x_d))]] \\
\end{align*}

As we use uniform distribution to choose the dropout channel on $C$ channels input, each individual channel $c$ is kept with probability $p(\delta^{(c)}\neq0)= \frac{C-1}{C}$. So we have
\begin{align*}
\mathbb{E}[f^-_{0}(x_d)] & = \frac{C-1}{C} \mathbb{E}[\sum_{c \in \mathcal{C}}W_0^{c}(\hat{f}_{0}^c(x_d))]] \\
& = \frac{C-1}{C} \mathbb{E}[f_{0}(x_d)]  
\end{align*}

To co-adapt the $W_0^{c}(\hat{f}_{0}^c(x_d))$ from different channels, we use adversarial loss on $f_{0}(x_d)$ and $\hat{f}_{0}(x_d)$  after compensation for the coefficient on their expectation's ratio, we use the following minimax adversarial loss to regularize the two distributions.
\begin{equation}
\mathcal{L}_{similarity} = \min_{\theta}\max_{\phi} \mathbb{E}_{x_d^{(\mathcal{C}_d^-)}}[\log D_\theta(\hat{f}_{0}(x_d,\phi))] + \mathbb{E}_{x_d^{(\mathcal{C}_d)}} [\log(1- D_\theta(\frac{C-1}{C}f_{0}(x_d,\phi))], 
\end{equation} 
where $D_\theta(\cdot)$ is an auxiliary discriminative network that are co-trained with our segmentation network.
The total loss function of our training is
\begin{equation}
\mathcal{L} = \mathcal{L}_{seg\_full} + \alpha \mathcal{L}_{seg\_drop} + \beta \mathcal{L}_{similarity}
\end{equation}

\subsection{Prediction Relevance Analysis}
\label{subsection:PediRele}
Robnik-Sikonja \etal \cite{robnik2008explaining} proposed a technique for assigning relevance value to each input feature with respect to a class label. Inspired by their idea, we assign a relevance value to each input channel $c$ with respect to segmentation label $j$ for every pixel. The basic idea is that the relevance of a segmentation label of a channel $j$ can be estimated by measuring how the prediction changes if we remove channel $c$.

We first estimate probability for class $j$ in pixel $\hat{y}_d$ produced from full channels inputs, denoted as $ \log p_Y( \hat{y}_d= j |x_d^{(\mathcal{C})})$. Then we re-estimate the class probability from missing channel $c$, denoted as $ \log p_Y( \hat{y}_d= j |x_d^{(\mathcal{C}^-_c)})$. Once these class probabilities are estimated, we follow the definition proposed by Robnik-Sikonja \etal \cite{robnik2008explaining} to calculate the weight of evidence on class $j$ from channel $c$, given as
\begin{equation}
\mathrm{WE}_c(y_d^j =c|x_d^c) = \log_2(\mathrm{odds}( p_Y( \hat{y}_d= j |x_d^{(\mathcal{C})}))) -\log_2(\mathrm{odds}( p_Y( \hat{y}_d= j |x_d^{(\mathcal{C}^-_c)}))),
\end{equation}
where $\mathrm{odds}(a) = a / (1-a) $.

%% file: subtex/experiment1.tex
\section{Experiments}
\paragraph{\bf Data, Data Selection and Preprocessing} We evaluate our method on BraTS17. BraTS17 contains 262 scans of glioblastoma and 199 scans of lower grade glioma. Our U-Net is trained on extracted 2D image patches which are extracted from all the 3D scans that have more than 200 non-zeros pixels (excluding the wasteful empty scans).  We randomly split the extracted 3D volume data to $80\%$ training and $20\%$ testing. We normalize the gray value to a range of -1 to 1 for each input channel and crop the MR image on the center window of size $200\times186$. We follow the protocol in BraTS17 to label the four types of intra-tumoral structures, namely GD-enhancing tumor (ET-label 4), the peritumoral edema (ED-label 2), the necrotic and non-enhancing tumor (NCR/NET label 1) and the background (label 0). For MR image scans, all the ground truth labels have been manually labeled and certified by expert board-certified neuroradiologists. The manual labeling process is followed by a hierarchical annotation protocol. Experts firstly segment whole tumor regions by contrasting on T2 and Flair channel and separate tumor core and enhancing core from T1 and T1c channel. Experts decision are agreed by majority vote.

\paragraph{\bf Evaluation} All our experiments were performed using Python 3.6 with TensorFlow library \cite{abadi2016tensorflow} and run on a GTX 1080 Ti graphics processing unit using CUDA 8.0. Our models are tested on both full channel input and missing channel input. In the case of missing channel input, each time we remove one channel from full channel input. Specifically, we ran the following experiments.

Firstly, we evaluated our segmentation result quantitatively by measuring Dice score with ground-truth. We report three types of Dice scores including whole tumor, tumor core and enhancing core on 2D slice level. Our three types of Dice score were evaluated on both full channel inputs and missing channel inputs.  

Secondly, we explored our segmentation results qualitatively by visualizing our results with ground truth segmentation. Our quality assessments include the segmentation region, boundary qualities and specificity/sensitivity. We evaluated our trained network's segmentation qualities on both full channel inputs and missing channel inputs.

Finally, to provide interpretations on which our prediction is based, we generated relevance map by the method we described in Section \ref{subsection:PediRele}. This helps us to understand the role of each individual channel on the estimation of whole tumor, tumor core and enhanced core. Our model has a surplus benefit of generating meaningful interpretations as our network broadly adapts to full channel and missing channel input.
\begin{figure}[htb!]
	\centering
	\includegraphics[width=1\linewidth]{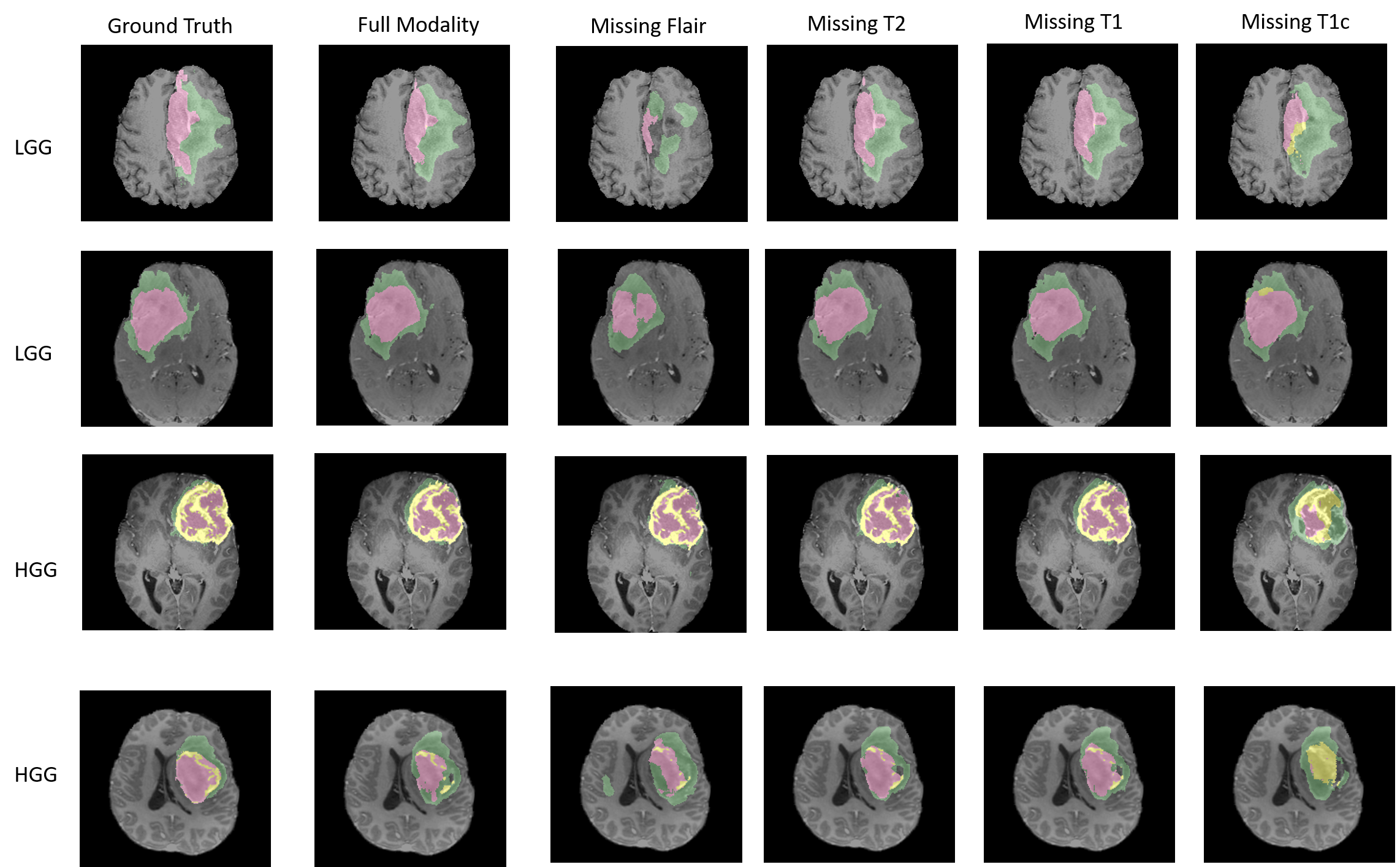}
	\caption{Examples of the segmentation masks with missing modality compared to the ground truth. The edema area is labeled in green color. The non-enhancing core area is labeled in red color. The enhancing core area is labeled in yellow color. } \label{fig:seg_exp_full}
\end{figure}
\paragraph{\bf Quantitative Result of our Model}
The segmentation result with our proposed method is shown in Table \ref{table:dice}. In the presence of four channels, our model can be comparable with the state-of-art result in \cite{jin2018ra}. The down-grading on Dice score for three types of tumor regions from missing channel inputs complies with the result in \cite{havaei2016hemis}. We also have the following findings: 
\begin{itemize}
    \item Missing T1 and T2 channel would have a minor decreasing in Dice score of all three categories.
    This observation consists with the manual protocol of BraTs dataset. T1 verifies the tumor core segmentation from T1c channel and T2 verifies whole tumor segmentation from Flair.
    
    \item Missing T1c channel would have a substantial decreasing in Dice score for both tumor core and enhancing core, for T1c is the chief indicating channel for segmenting tumor core and enhancing core region. Missing Flair channel would have a sharp decreasing on Dice score for all three categories, for Flair is the chief indicating channel for segmenting the whole tumor.
    
    \item As shown in Fig. \ref{fig:seg_exp_full}, missing Flair channel would result in a coarse locating on the whole tumor region. And missing T1c channel would result in mistaking tumor core and enhanced core region as non-core region.
    
\end{itemize}

To provide justification of domain adaptation loss term we introduce, we also provide an ablation experiment by training only on reconstruction loss in training process. Our baseline result is shown in Table \ref{table:dice}. Distinct increases of Dice score in all categories are observed when trained with domain adaptation loss compared with the baseline. The qualitative visualization are consistent with the quantitative results as shown in Fig. \ref{fig:seg_exp_full}. 
\begin{table}
	\centering
	\setlength{\tabcolsep}{10pt}
	\begin{tabular} {| c | c | c | c | c | c | c |}
		\hline
		{} &  \multicolumn{3}{|c|}{\textbf{Proposed}} &  \multicolumn{3}{|c|}{\textbf{Baseline}} \\ \hline
		&  WT &  TC &  EC &  WT &  TC &  EC
		\\ \hline
		Full channel  &0.894  & 0.790  & 0.653  & 0.875  &  0.693 & 0.554
		\\ \hline
		Missing T1  & 0.890   &  0.778  &  0.642  & 0.871  &  0.672  & 0.532 
		\\ \hline
		Missing T1c& 0.879   &  0.570  &  0.484 &  0.856  & 0.426  &  0.380
		\\ \hline
		Missing T2  & 0.893  &  0.775  &  0.643 & 0.865  & 0.660   &  0.521
		\\ \hline
		Missing Flair  & 0.616  & 0.680  &  0.552 & 0.508  & 0.570  &  0.464
		\\ \hline	
	\end{tabular}
	\caption{The dice scores for whole tumor (WT), tumor core (TC) and enhanced core (EC) on the test dataset. The left is the result of our proposed method with our adversarial loss. The right is result of our baseline method trained without our domain adaptation loss.} \label{table:dice}
\end{table}
\paragraph{\bf Contribution of Every Channel to Segmentation Labels}
In contrast to most segmentation algorithms performed as black-boxes and evaluated purely on accuracy, we provide some insights of the decisions of segmentation networks to physicians. The physicians could weigh this information and incorporate it in the overall diagnosis process.
\begin{figure}[htb!]
	\centering
	\includegraphics[width=1\linewidth]{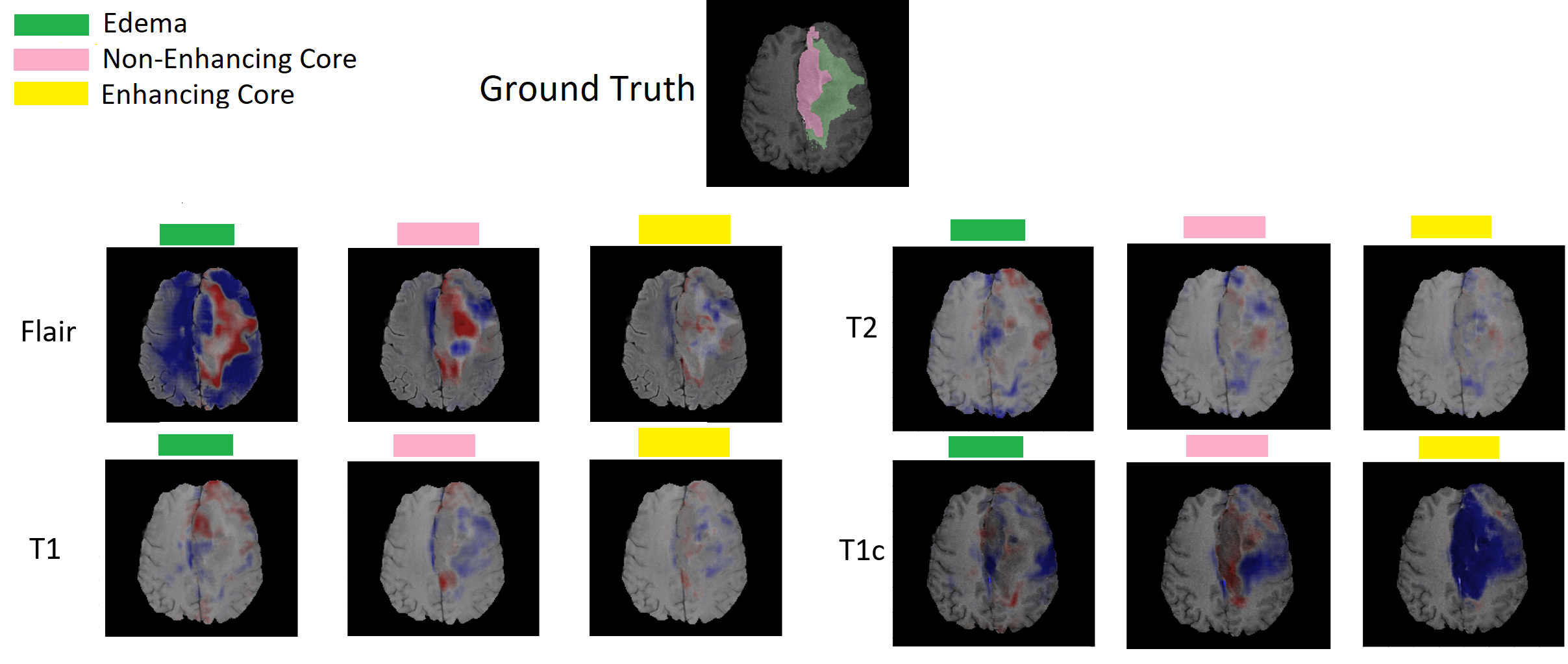} 
	\caption{Visualization of segmentation decision on different channels, using prediction relevance analysis. Red color represents positive evidence, and blue color represents negative evidence of that region. } \label{fig:seg_interp}
\end{figure}

Fig. \ref{fig:seg_interp} shows the contribution of each channel while predicting three types of tumors at pixel level. By weighting the prediction differences between inputting full channels and removing a specific channel, we visualize the influence of each channel to an individual segmentation label. Red color represents positive evidence, and blue color represents negative evidence from that channel to the segmentation label. We observed that, different channels contribute distinctly to different tumor types. 

For example, Flair channel provides a strong positive evidence on the whole tumor regions and a negative evidence of the non-tumor regions. T2 channel refines the evidence across the boundary area. T1c channel provides supporting evidence both on tumor core and enhanced core area against the rest in the whole tumor area. T1 channel refines the evidence across the boundaries of tumor core area. In general, we trust the segmentation classifiers that could not only produce exact and distinct labels for each tumor type, but also provide reasonable explanations for its decisions.   

%% file: subtex/conclusion1.tex
%
%

\section{Conclusion}
We propose a brain tumor segmentation algorithm that is robust to missing modality. Our model is designed to recover the information from missing modality and is able to visualize the contribution of each channel. The comparisons between full and missing modality show the important roles of Flair and T1c on discrimination of whole tumor and tumor core, respectively. These findings are along well with the expert labeling routine \cite{menze2015multimodal}.